\documentclass{article}
\usepackage{ijcai19}

\usepackage{times}
\usepackage{soul}
\usepackage{url}
\usepackage[hidelinks]{hyperref}
\usepackage[utf8]{inputenc}
\usepackage[small]{caption}
\usepackage{graphicx}
\usepackage{amsmath}
\usepackage{booktabs}
\usepackage{algorithm}
\usepackage{algorithmic}
\usepackage{amsfonts}
\usepackage{multirow}
\title{Legal Judgment Prediction via Multi-Perspective Bi-Feedback Network}

\author{
Wenmian Yang$^{1,2}$
\and
Weijia Jia*$^{2,1}$\and
Xiaojie Zhou$^1$\And
Yutao Luo$^1$
\affiliations
$^1$Department of Computer Science and Engineering, Shanghai Jiao Tong University, China\\
$^2$State Key Lab of IoT for Smart City, CIS, University of Macau, Macao, SAR China\
\emails
\{sdq11111, szxjzhou,luoyt1996\}@sjtu.edu.cn,
jiawj@um.edu.mo,
}

\begin{document}

\maketitle

\begin{abstract}
The Legal Judgment Prediction (LJP) is to determine judgment results based on the fact descriptions of the cases. LJP usually consists of multiple subtasks, such as applicable law articles prediction, charges prediction, and the term of the penalty prediction. These multiple subtasks have topological dependencies, the results of which affect and verify each other. However, existing methods use dependencies of results among multiple subtasks inefficiently. Moreover, for cases with similar descriptions but different penalties, current methods cannot predict accurately because the word collocation information is ignored. In this paper, we propose a Multi-Perspective Bi-Feedback Network with the Word Collocation Attention mechanism based on the topology structure among subtasks. Specifically, we design a multi-perspective forward prediction and backward verification framework to utilize result dependencies among multiple subtasks effectively. To distinguish cases with similar descriptions but different penalties, we integrate word collocations features of fact descriptions into the network via an attention mechanism. The experimental results show our model achieves significant improvements over baselines on all prediction tasks.
\end{abstract}

\section{Introduction}
\label{intro}
The task of legal judgment prediction (LJP) is to determine the judgment results based on the fact descriptions of the cases. Such techniques are crucial for legal assistant systems, which are helpful for ordinary people who are unfamiliar with legal terminology and complex procedures in legal consulting. Moreover, it can also serve as the handy reference for professionals (e.g., lawyers and judges) and improve their work efficiency. 

So far, LJP has been regarded as a classic text classification problem and studied for many years \cite{segal1984predicting,lauderdale2012supreme,ye2018interpretable,luo2017learning}. However, there still exist two major challenges that make existing methods not performed well as detailed below:

\textbf{Topological Dependencies among Multi-task Results}: Generally, LJP consists of multiple subtasks, such as the charge prediction and the term of the penalty prediction \cite{xiao2018cail2018}. Moreover, \cite{zhong2018legal} consider that the law article prediction is also one of the fundamental subtasks in countries with the civil law system (e.g., China and Germany), and these multiple subtasks have a strict order in the real world. Specifically, appropriate law articles are decided by human judges at first, and then charges are determined according to the law articles. Finally, the term of the penalty is further confirmed based on laws and charges. \cite{zhong2018legal} explore the topological dependencies of LJP multiple subtasks according to the above order, and propose a topological multi-task learning framework. However, their work ignores the interaction of task results. Most of the multi-task learning methods ignore the dependencies among multi-task results too, since they generally focus on sharing representations (or some encoding layers) among relevant tasks \cite{liu2015representation} or sharing hard/soft parameters \cite{zhang2017survey}. In some scenarios where multitasking exists topology dependencies as LJP, the prediction results of each task can also affect and verify its related tasks. For example, if the applicable laws are predicted as property infringement, then the content related to property infringement in the fact description should be emphasized when predicting charges. Meanwhile, the prediction of the charges can also verify the rationality of predicted applicable laws. Therefore, it is a major and meaningful challenge to exploit the result of each task to infer and validate other tasks based on the topological dependencies.

\textbf{Cases of Similar Descriptions with Different Penalties}: In LJP, some cases have similar fact descriptions with different judgments. \cite{hu2018few} manually introduce several discriminative attributes of charges as the internal mapping between fact descriptions and charges, which provide effective signals for distinguishing confusing charges (also suitable for laws). However, the term of penalty is difficult to distinguish with limited attributes. For example, the cases ``Zhang smuggled gold 10kg and drugs 10g'', and ``Zhang smuggled drugs 10kg and gold 10g'' have the same charge, but completely different penalties in the real world.  Most existing work cannot distinguish them, which is mainly because the word collocation information in the fact description is ignored. Moreover, the semantics associated with numbers are not well interpreted. Existing natural language techniques often ignore the semantics of numbers because numbers are not crucial in general problems. However, when predicting the term of the penalty, the semantics of number in the text is quite crucial. Therefore, how to utilize word collocations effectively and extract the semantics of numbers to distinguish cases with similar descriptions but different penalties is also a major challenge. 

To solve the above challenges, in this paper, we propose a Multi-Perspective Bi-Feedback Network (MPBFN) with Word Collocation Attention (WCA) mechanism. To incorporate the predicted results of subtasks into the network, we design the MPBFN with multi-perspective Forward Prediction (FP) and Backward Verification (BV), i.e., bi-feedback, which improves the overall performance of LJP.  Specifically, we first map the intermediate prediction result of each task into the latent state space. Then we merge the latent state with the semantic vector of fact and obtain the task result-based semantic vector, which is utilized to make the FP for its follow-up tasks. Meanwhile, we also make a BV to check the rationality of its pre-order tasks based on the latent state. Finally, for each task, we combine all prediction and verification results, which we call multi-perspective (MP), to obtain the final prediction results. To better distinguish cases with similar descriptions but different penalties, we extract the collocation information and the semantics of numbers with attention mechanism. Specifically, we first consider the number as a digit string and embed it by digit to obtain the number semantic vector. Then, we extract the collocations in the fact description and obtain the weights of collocations by result-based attention mechanism. Finally, we combine weighted collocation features with MPBFN to optimize the performance of the penalty prediction.

The main contributions of this paper are presented as follows:
\begin{enumerate}
\item We are the first to focus on and utilize the dependencies among prediction results of multiple subtasks in LJP and propose the MPBFN.
\item We extract the semantics of numbers and utilize the word collocation information in the fact description, and propose a WCA mechanism to distinguish cases with similar descriptions but different penalties.
\item We evaluate the MPBFN-WCA with two real-world datasets, and our model significantly outperforms all baselines on all subtasks.

\end{enumerate}

\section{Related Work}
In this section, we discuss the related work in two aspects.

\textbf{Legal Judgment Prediction} has been studied for decades and most existing work formalizes this task under the text classification framework \cite{kort1957predicting,segal1984predicting,lauderdale2012supreme}. Since machine learning has been proven successful in many areas, researchers begin to formalize LJP with machine learning methods. \cite{sulea2017exploring} develop an ensemble system, which averages the output of multiple SVM classifiers to improve the performance of LJP related text classification. Recently, neural network-based methods have been applied to LJP and have achieved better results than traditional machine learning methods. \cite{luo2017learning} propose an attention-based neural network method to jointly model the charge prediction task and the relevant article extraction task in a unified framework, which solves multi-label tasks in LJP. Besides, \cite{ye2018interpretable} explore charge labels to solve the non-distinctions of fact descriptions and adopt a Seq2Seq model to generate court views and predicted charges. \cite{hu2018few} propose an attribute-attentive charge prediction model to infer the attributes and charges simultaneously, which is the first to focus on few-shot and confusing problems. Moreover, legal judgment usually consists of complicated subclauses, and there exists a strict order among the subtasks of legal judgment. \cite{zhong2018legal} first explore and formalize the multiple subtasks of legal judgment, and propose a topological multi-task learning framework. However, their work ignores the interaction of task results, which is quite meaningful in LJP.

\textbf{Multi-task Learning} has numerous successful usages in NLP tasks. Sharing representations (or some encoding layers) among relevant tasks \cite{liu2015representation} or sharing hard/soft parameters \cite{zhang2017survey,D18-1243} are two general ways. For instance, \cite{luong2015multi} share encoders or decoders to improve one (many) to many neural machine translation. \cite{P18-1064} improve abstractive summarization via multi-task learning with the auxiliary tasks of question generation and entailment generation.  \cite{P18-1074} propose a multi-lingual multi-task architecture to develop supervised models with a minimal amount of labeled data for sequence labeling. Different from existing methods, in this paper, we utilize the dependencies between the prediction results of multiple subtasks and proposed a Multi-Perspective based Bi-Feedback Network.

\section{Method}
In this section, we describe the Multi-Perspective based Bi-Feedback Network (MPBFN) and a Word Collocation Attention (WCA) mechanism. Specifically, the problem formulation and overview are first provided in Section \ref{PF} and Section \ref{OV}, respectively. Then, we provide a neural encoder for fact descriptions in Section \ref{encoder}. Next, we introduce the MPBFN and a WCA mechanism in Section \ref{MPBFN} and \ref{WCA}, respectively. The training process of the model is finally introduced in Section \ref{train}.

\subsection{Problem Formulation}
\label{PF}
For each case, suppose the fact description is a word sequence $\textbf{A}=\{a_1,...,a_l\}$, where $l$ is the number of words. Moreover, we extract $n$ word collocations from the fact description by Stanford CoreNLP toolkit \cite{manning-EtAl:2014:P14-5}, and obtain the word collocation sequence $\textbf{COL}=\{\textbf{col}_1,...,\textbf{col}_{n}\}$. Given $\textbf{A}$ and $\textbf{COL}$ as input, our task is to predict the judgment results of applicable law articles, charges and term of penalty, which is a multi-task classification problem.

\subsection{Overview}
\label{OV}
Our MPBFN-WCA consists of three parts, i.e., the encoder, MPBFN, and WCA mechanism. In this paper, we employ the CNN encoder [Kim, 2014] to generate the semantic vector of the fact description. 
\begin{figure}[htbp]
  \centering
  \includegraphics[width=0.75\linewidth]{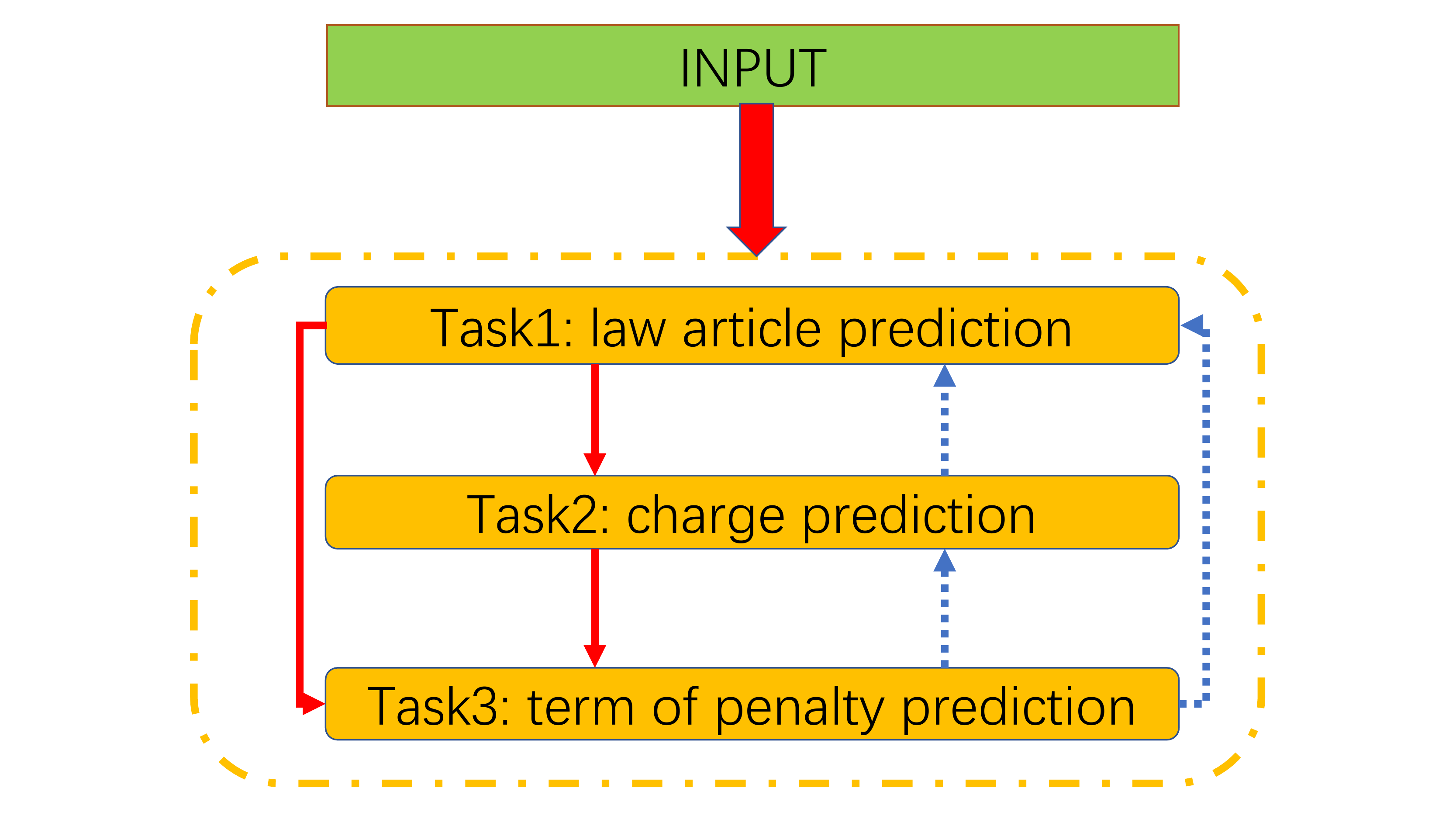}
  \caption{Topology of LJP tasks}\label{graph}
\end{figure}

After getting the encoder, we design the MPBFN as the decoder. Before proposing MPBFN, we first introduce the topology dependencies among subtasks in LJP. As mentioned in Section \ref{intro}, there exists a strict order among subtasks of LJP \cite{zhong2018legal}, which is shown in Figure \ref{graph}. To facilitate the following calculations, we define the law article prediction, charge prediction, and penalty term prediction as task 1, task 2, and task 3, respectively.

Based on the topology, we design the MPBFN to utilize the dependencies among prediction results of subtasks in LJP. In MPBFN, the result of each task can infer its follow-up tasks, and can also verify its pre-order tasks. For instance, according to the law article prediction (task 1), we emphasize the content related to applicable laws in the fact description and make a more attentive FP for the charge (task 2), denoted as FP (1, 2). Conversely, based on the charge prediction, we make a BV$(2, 1)$ to check whether the law articles are applicable. That is the philosophy of bi-feedback. In Figure \ref{graph}, the solid red lines express the FP, and the dotted blue lines express the BV. Besides, a task may have more than one follow-up task and pre-order task. For each task, different pre-order/follow-up tasks may have different prediction/verification results because different subtasks pay attention to different parts of the fact description. Combining the results based on different related tasks can obtain more accurate results. That is what we call the MP. 
\begin{figure}[htbp]
  \centering
  \includegraphics[width=0.95\linewidth]{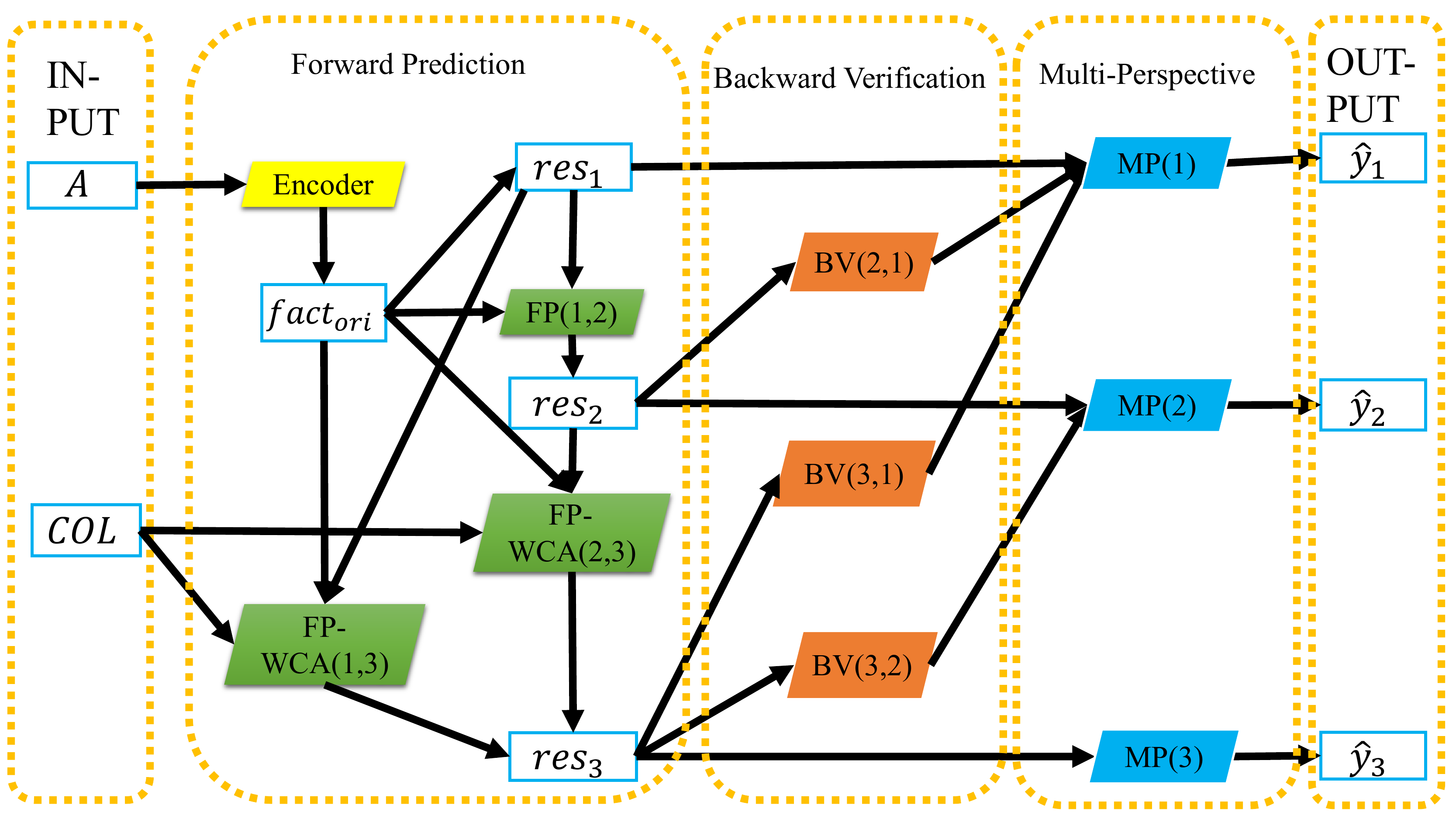}
  \caption{General framework of MPBFN-WCA}\label{network}
\end{figure}

Finally, to distinguish cases with similar descriptions but different penalties, we propose the WCA mechanism. We extract the semantic features of numbers and collocations in the fact description and combine them with MPBFN to predict task 3 more accurately. 

The general framework of MPBFN-WCA is shown in Figure \ref{network}, where $\textbf{fact}_{ori}$ expresses the original fact semantic vector, $\textbf{res}_i$ is comprehensive FP result of task $i$, and $\hat{y_i}$ denotes the final prediction result of task $i$.

\subsection{Neural Encoder for Fact Descriptions}
\label{encoder}
In this section, we employ the CNN encoder \cite{kim2014convolutional} to generate the semantic vector of the fact description.

Given a word sequence $\textbf{A}=\{a_1,...,a_l\}$, we first convert each word $a_i$ into its word embedding $\textbf{x}_i \in \mathbb{R}^{d_w}$, and get the word embedding sequence $\textbf{X}=\{\textbf{x}_1,...,\textbf{x}_l\}$, where $d_w$ is the dimension of the word embedding.

For $\textbf{X}$, we make a convolution operation with the convolution martix $\textbf{W}_c\in \mathbb{R}^{d_c\times (h\times d_w)}$ by 
\begin{equation}
\textbf{c}_i=\textbf{W}_c\cdot \textbf{x}_{i:i+h-1} + \textbf{b}_c, i \in [1,l-h+1]
\end{equation}
where $\textbf{x}_{i:i+h-1}$ is the concatenation of word embeddings in the $i$-th window, $\textbf{b}_c\in \mathbb{R}^{d_c}$ is the the bias vector, $d_c$ is the number of filters and $h$ is the length of a sliding window.
We apply the convolution over each window $i$ and get ${\textbf{C}=\{\textbf{c}_1,\textbf{c}_2,...,\textbf{c}_{l-h+1}}\}$.

Finally, we apply the max-pooling over $\textbf{C}$ by
\begin{equation}
f_k=max(c_{1,k},c_{2,k},...,c_{l-h+1,k}), k\in [1,d_c]
\end{equation}
and obtain the fact semantic vector 
\begin{equation}
\textbf{fact}_{ori} = \{f_1,f_2,...,f_{d_c}\}
\end{equation}
\subsection{Multi-Perspective based Bi-Feedback Network}
\label{MPBFN}
In this section, we first introduce the FP and BV, respectively. Then, we calculate the final result of each task by combining results of multi-perspective FP and BV.

\subsubsection{Forward Prediction}
In LJP, each category of a subtask result has a specific meaning. For example, in the law article prediction task, each category of the result represents a specific law article. Therefore, for task $i$, we employ a latent state matrix $\textbf{S}^i \in \mathbb{R}^{t_i\times d_s}$ (an embedding layer) to express the latent state of the result, where $t_i$ is the number of result categories of task $i$, and $d_s$ is the dimension of the latent state.

As shown in Figure \ref{network}, we build the network according to the topology. Assume that we have predicted the result distribution of task $i$, $\textbf{res}_i \in \mathbb{R}^{t_i}$ based on its pre-order tasks (see Section ``Multi-Perspective Integration'' for details). We combine the result distribution $\textbf{res}_i$ with the latent state matrix $\textbf{S}^i$, and obtain the result-based latent state vector $\textbf{lsv}_i \in \mathbb{R}^{d_s}$ by
\begin{equation}
\label{lsv}
\textbf{lsv}_i = \sum_{j=1}^{t_i}{res_{i,j} \cdot \textbf{S}_{j}^{i}}
\end{equation}
where $j$ expresses the $j$-th category of the result of task $i$.

Then, we map the latent state vector $\textbf{lsv}_i$ to the corresponding semantic space $\textbf{sem}_i \in \mathbb{R}^{d_c}$ by 
\begin{equation}
\label{sem}
\textbf{sem}_i = elu(\textbf{Ws}_i\cdot \textbf{lsv}_i)
\end{equation}
where $\textbf{Ws}_i \in \mathbb{R}^{d_c \times d_s}$ is the transformation matrix from latent state space to semantic space of task $i$ and $elu$ is an activation function proposed by \cite{clevert2015fast}.

We merge the result-based semantic vector $\textbf{sem}_i$ with the semantic vector of the fact description $\textbf{fact}_{ori}$, and obtain the task $i$ based fact semantic vector by
\begin{equation}
\label{fact}
\textbf{fact}_i = \textbf{fact}_{ori}\otimes \textbf{sem}_i
\end{equation}
where $\otimes$ is an element-wise product function. 

Finally, we predict each follow-up task $j$ of task $i$ by
\begin{equation}
\label{pred}
\textbf{pred}_{i,j} = softmax(\textbf{Wf}_{i,j}\cdot \textbf{fact}_i + \textbf{bf}_{i,j})
\end{equation}
where $\textbf{Wf}_{i,j} \in \mathbb{R}^{t_j \times d_c}$ is the fully connected matrix and $\textbf{bf}_{i,j}\in \mathbb{R}^{t_j}$ is the bias vector.

\subsubsection{Backward Verification}
Conversely, the result of each task can verify pre-order tasks.

For task $i$, we first calculate the result-based latent state vector $\textbf{lsv}_i$ by Eq. (\ref{lsv}). Then we map $\textbf{lsv}_i$ to the gate vector space of its pre-order task $j$, and obtain
$\textbf{gate}_{i,j} \in \mathbb{R}^{t_j}$
by
\begin{equation}
\textbf{gate}_{i,j} = sigmoid(\textbf{Wg}_{i,j}\cdot \textbf{lsv}_i + \textbf{bg}_{i,j})
\end{equation}
where $\textbf{Wg}_{i,j} \in \mathbb{R}^{t_j \times d_s}$ is the fully connected matrix and $\textbf{bg}_{i,j}\in \mathbb{R}^{t_j}$ is the bias vector.

We employ the sigmoid function to determine whether each category of the result of task $j$ is available based on the result of task $i$. That is why we call $\textbf{gate}_{i,j}$ the gate vector.

\subsubsection{Multi-Perspective Integration}
After we obtain FP and BV results, in this section, we merge these multi-perspective results and calculate the final result of each task.

First, we summarize multi-perspective FP results for each task in topological order, i.e., before solving task $i$, each task $j$ that $j<i$ in topological order has been solved. For task $i$, we calculate the comprehensive FP result $\textbf{res}_i \in \mathbb{R}^{t_i}$ by
\begin{equation}
\textbf{res}_i=\left\{
\begin{aligned}
softmax(\textbf{Wf}_{i}\cdot \textbf{fact}_{ori} + \textbf{bf}_{i}) && p_i = 0 \\
norm(\prod_{j=1}^{p_i}{\textbf{pred}_{j,i}}) && p_i > 0
\end{aligned}
\right.
\end{equation}
where $p_i$ is the number of pre-order tasks of task $i$ and $norm(x)$ is the normalization function. We take the intersection of each FP result and obtain a more accurate result.

Similarly, the comprehensive BV result of task $i$ $\textbf{ver}_i \in \mathbb{R}^{t_i}$ can be calculated by 
\begin{equation}
\textbf{ver}_i=\left\{
\begin{aligned}
e && u_i = 0 \\
norm(\prod_{j=1}^{u_i}{\textbf{gate}_{j,i}}) && u_i > 0
\end{aligned}
\right.
\end{equation}
where $u_i$ is the number of follow-up tasks of task $i$ and $e$ is the vector with 1 in each dimension.

Finally, we merge the results of comprehensive FP and BV of task $i$ by
\begin{equation}
\hat{y_i} = \textbf{res}_i \otimes \textbf{ver}_i
\end{equation}
which is the final prediction result of task $i$.

\subsection{Word Collocation Attention mechanism}
\label{WCA}
In this section, we propose a number embedding method and a word collocation attention mechanism to increase the accuracy of the penalty prediction in cases where the fact description is similar. 
\subsubsection{Number Embedding}
In the penalty prediction, numerals with their units have a significant impact. In this section, we propose a digital bitwise-based embedding method to obtain the number embedding.

In particular, we first convert a number $i$ into a fixed $ln$-bit vector $\textbf{dig}_i=\{dig_{i,1},dig_{i,2},...,dig_{i,ln}\}$. The unit digit corresponds to $dig_1$, the tens digit corresponds to $dig_2$, and so forth. We pad zeros when number $i$ is less than $ln$-bit and discard high digits when exceeds.

Then, we map each digit into the semantic space by embedding layer, and obtain 
\begin{equation}
\textbf{num}_i=\textbf{num}_{i,1}\oplus \textbf{num}_{i,2}...\oplus\textbf{num}_{i,ln}
\end{equation}
where $\textbf{num}_{i,j} \in \mathbb{R}^{d_n}$ is the embedding vector of $dig_{i,j}$ and we assure that $d_n \times ln = d_c$.

Finally, we merge the number semantic vector $\textbf{num}_i$ with its corresponding unit's word embedding vector $\textbf{x}_k$, and obtain:
\begin{equation}
\label{ns}
\textbf{ns}_i = tanh(\textbf{Wn}\cdot (\textbf{num}_i \oplus \textbf{x}_k) + \textbf{bn}))
\end{equation}
where $k$ is the position of the unit in the word sequence $\textbf{A}$ (labeled by Stanford CoreNLP toolkit \cite{manning-EtAl:2014:P14-5}),
$\textbf{Wn} \in \mathbb{R}^{d_c \times 2d_c}$ is the fully connected matrix and $\textbf{bn}\in \mathbb{R}^{d_c}$ is the bias vector.

Since the semantics of numerals is mainly used for the penalty prediction, we do not utilize the number embedding directly in the fact description encoder, but in the collocation embedding that will describe in the next section.

\subsubsection{Collocation Attention}
With the semantics of numbers, we further focus on the impact of word collocation on the penalty prediction in this section. 

In the word collocation sequence $\textbf{COL}=\{\textbf{col}_1,...,\textbf{col}_{n}\}$, each collocation $\textbf{col}_i=(w_{i,1},w_{i,2})$ contains two words. We convert each word to its corresponding  semantic vector and obtain the collocation embedding $\textbf{ecol}_i=(\textbf{ew}_{i,1},\textbf{ew}_{i,2})$ and the sequence $\textbf{ECOL}=\{\textbf{ecol}_1,...,\textbf{ecol}_n\}$, where $\textbf{ew}_{i,j}$ is the word embedding vector of word $w_{i,j}$. Particularly, if a word in the collocation is a number with its unit, we replace the original word embedding vector with the vector that calculated by Eq. (\ref{ns}).

Then, we merge the word vector $\textbf{ew}_{i,1}$ with $\textbf{ew}_{i,2}$ and obtain the collocation semantic vector $\textbf{ec}_i \in \mathbb{R}^{d_c}$ by LSTM \cite{hochreiter1997long}
\begin{eqnarray}
&&\textbf{h}_1 = LSTM(\textbf{ew}_{i,1})\\
&& \textbf{h}_2 = LSTM(\textbf{ew}_{i,2},\textbf{h}_1)\\
&&\textbf{ec}_i = (\textbf{h}_1 + \textbf{h}_2) / 2
\end{eqnarray}
where $\textbf{h}_j$ is the hidden state vector of LSTM, and obtain the collocation semantic vector sequence $\textbf{EC}=\{\textbf{ec}_1,...,\textbf{ec}_n\}$.

Different word combinations have different influences on different task results. Therefore, for each task $j$, we design an attention mechanism to obtain a result-based collocation semantic vector $\textbf{ect}_j \in \mathbb{R}^{d_c}$ by
\begin{equation}
\textbf{ect}_j = \sum_{i=1}^{n}{\alpha_{j,i}\cdot \textbf{ec}_i}
\end{equation}
where $\alpha_{j,i}$ is the attention values of collocation $i$ based on task $j$ that can be calculated by 
\begin{equation}
\alpha_{j,i}= \frac{exp(tanh(\textbf{Wa}_{j} \cdot \textbf{ec}_i)^{T}\cdot \textbf{sem}_j)}{\sum_{k}{exp(tanh(\textbf{Wa}_{j} \cdot \textbf{ec}_k)^{T}\cdot \textbf{sem}_j)}}
\end{equation}
The $\textbf{Wa}_j \in \mathbb{R}^{d_c \times d_c}$  is the a weight matrix and $\textbf{sem}_j$ is calculated by Eq. (\ref{sem}).

Finally, for task $j$, we merge its collocation-based semantic vector $\textbf{ect}_j$ with the fact semantic vector $\textbf{fact}_j$, and get the mixed semantic vector $\textbf{fc}_j$ by
\begin{equation}
\textbf{fc}_j = elu(\textbf{Wm}_j\cdot (\textbf{ect}_j \oplus \textbf{fact}_{ori}) + \textbf{bm}_j)\otimes \textbf{fact}_j
\end{equation}
where $\textbf{Wm} \in \mathbb{R}^{d_c \times 2d_c}$ is the fully connected matrix and $\textbf{bm}\in \mathbb{R}^{d_c}$ is the bias vector.

In this paper, we employ the mixed semantic vector to predict task 3, the term of penalty. Specifically, we replace the $\textbf{fact}_1$ and $\textbf{fact}_2$ with $\textbf{fc}_1$ and $\textbf{fc}_2$ when calculating $\textbf{pred}_{1,3}$ and $\textbf{pred}_{2,3}$ by Eq.(\ref{pred}). 

\subsection{Training}
\label{train}
In this paper, we assume that the prediction result of each task is single-label. We use cross-entropy loss function for each subtask and take the sum as an overall loss by:
\begin{equation}
Loss = -\sum_{k=1}^{3}{\sum_{j=1}^{t_k}{\textbf{y}_{k,j}log(\hat{\textbf{y}_{k,j}})}}
\end{equation}
We train our model in an end-to-end fashion, and use Adam \cite {kinga2015method} for optimization. Besides, we utilize the dropout after the fact semantic vector layer $\textbf{fact}_{ori}$ to prevent overfitting.

\section{Experiments}
In this section, we demonstrate the effectiveness of our model. We first introduce datasets and data processing. Then we provide the necessary parameters of our model. Finally, we compare the performance of our model with baselines and analyze the effect of each module in our model.
\subsection{Dataset Construction and Experimental Setup}
In this section, we use two public datasets from Chinese AI and Law challenge (CAIL2018) \cite{xiao2018cail2018}, \textit{i.e.}, CAIL-small (the exercise stage data) and CAIL-big (the first stage data). 

The cases in the datasets contain fact descriptions, applicable law articles, charges and the terms of penalty. Also, there exist cases with multiple applicable law articles and multiple charges. As our model aims to explore the effectiveness of considering result-based topological dependencies between various subtasks, we filter out these multi-label samples.

Moreover, there are also some infrequent charges and law articles, for example, insulting the national flag and national emblem. We filter out these infrequent charges and law articles and only keep those with frequencies greater than 100. For the term of penalty, we divide the terms into non-overlapping intervals as \cite{zhong2018legal}. The detailed description of the datasets is shown in Table \ref{dataset}. Particularly, there exists a validation set with 12,787 cases in CAIL-small.

\begin{table}[!htbp]\footnotesize
\centering
\caption{The description of datasets}
\label{dataset}
\begin{tabular}{c|c|c}
\toprule 
Datasets&CAIL-small&CAIL-big\\
\hline  
Training Set Cases &101,685&1,588,894\\
Test Set Cases&26,766&185,228\\
Law Articles&103&118\\
Charges&119&130\\
Term of Penalty &11&11\\
\bottomrule
\end{tabular}
\end{table}

Since the cases are written in Chinese, we use THULAC \cite{sun2016thulac} for word segmentation. For word embedding, we set the frequency threshold as 25, and initialize the word embedding vector with a uniform distribution directly. We extract the word collocations in the fact description by Stanford CoreNLP toolkit \cite{manning-EtAl:2014:P14-5}. Specifically, we employ the CoreNLP to extract the basic dependencies in each sentence of the case as collocations and set the maximum number of collocations in each case as 128. Particularly, we treat the number and its unit as one word. For CNN-based model in baselines, we set the maximum document length as 512 words; for LSTM-based model in baselines, we set the maximum sentence length to 64 words and maximum document length to 64 sentences. 
\begin{table*}[!htbp]\footnotesize
\centering
\caption{Judgment prediction results on CAIL-small.}
\label{cail-small}
\begin{tabular}{c|c|cccc|cccc|cccc}
\toprule 
&Tasks&\multicolumn{4}{|c|}{Law Articles} &\multicolumn{4}{|c|}{Charges} & \multicolumn{4}{|c}{Term of Penalty} \\
\hline
& Metrics & Acc. & MP & MR & F1 & Acc. & MP & MR & F1 & Acc. & MP & MR & F1 \\
\hline
\multirow{4}*{Baselines}& FLA & 0.803 & 0.724 & 0.720 & 0.714 & 0.767 & 0.758 & 0.738 & 0.732 & 0.371 & 0.310 & 0.300 & 0.299 \\
~& HARNN & 0.852 & 0.785 & 0.773 & 0.775 & 0.857 & 0.837 & 0.821& 0.826 & 0.392 & 0.365 & 0.346 & 0.352 \\
~& CNN & 0.866 & 0.811 & 0.782 & 0.788 & 0.854 & 0.857 & 0.835& 0.837 & 0.392& 0.362 & 0.342 & 0.339 \\
~& TOPJUDGE & 0.872& 0.819 & 0.808 & 0.800 & 0.871 & 0.864 & 0.851 & 0.846 & 0.380& 0.350 & 0.353 & 0.346 \\
\hline
\multirow{3}*{Ours}& MPFP & 0.870 & 0.821 & 0.805 & 0.797 & 0.875 & 0.862 & 0.841 & 0.843 & 0.405 & 0.390 & 0.369 & 0.377 \\
~&MPBFN& 0.878 & 0.828& 0.814 & 0.815 & 0.883 & 0.871 & 0.851 & 0.849 & 0.404 & 0.387 & 0.358 & 0.374 \\
~& \textbf{MPBFN-WCA} & \textbf{0.883} & \textbf{0.832} & \textbf{0.824} & \textbf{0.822} & \textbf{0.887}& \textbf{0.875} & \textbf{0.857} & \textbf{0.859} & \textbf{0.414 }& \textbf{0.406} & \textbf{0.369 }& \textbf{0.392} \\
\bottomrule
\end{tabular}
\end{table*}

\begin{table*}[!htbp]\footnotesize
\centering
\caption{Judgment prediction results on CAIL-big.}
\label{cail-big}
\begin{tabular}{c|c|cccc|cccc|cccc}
\toprule 
&Tasks&\multicolumn{4}{|c|}{Law Articles} &\multicolumn{4}{|c|}{Charges} & \multicolumn{4}{|c}{Term of Penalty} \\
\hline
& Metrics & Acc. & MP & MR & F1 & Acc. & MP & MR & F1 & Acc. & MP & MR & F1 \\
\hline
\multirow{4}*{Baselines}& FLA & 0.942 & 0.763 & 0.695 & 0.746 & 0.931 & 0.798 & 0.747 & 0.780 & 0.531 & 0.437& 0.331 & 0.370 \\
~& HARNN & 0.951& 0.870 & 0.733 & 0.784 & 0.944& 0.881 & 0.794 & 0.826 & 0.575 & 0.494& 0.417 & 0.433 \\
~& CNN & 0.955 & 0.861 & 0.754 & 0.789 & 0.954 & 0.896 & 0.805& 0.838 & 0.569 & 0.482 & 0.396& 0.422 \\
~& TOPJUDGE & 0.963 & 0.870 & 0.778 & 0.802& 0.960 & 0.906 & 0.824 & 0.853 & 0.569& 0.480 & 0.398 & 0.426 \\
\hline
\multirow{3}*{Ours}& MPFP & 0.958 & 0.861 & 0.775 & 0.797 & 0.961 & 0.900 & 0.824 & 0.848 & 0.579 & 0.516 & 0.408 & 0.446 \\
~&MPBFN& 0.964& 0.864 & 0.782 & 0.812 & 0.971 & 0.908 & 0.832 & 0.860 & 0.583 & 0.516 & 0.412& 0.450\\
~& \textbf{MPBFN-WCA} & \textbf{0.978} & \textbf{0.872} & \textbf{0.789} & \textbf{0.820} & \textbf{0.977}& \textbf{0.914} & \textbf{0.836} & \textbf{0.867} & \textbf{0.604}& \textbf{0.534} & \textbf{0.430 }& \textbf{0.464} \\
\bottomrule
\end{tabular}
\end{table*}

Meanwhile, we use Keras (https://keras.io/) framework to build neural networks. We set the word embedding size $d_w$ as 200, and the dimension of the latent state $d_s$ as 256. For CNN encoder, we set the number of filters $d_c$ as 256, and the length of sliding window $h$ as 2,3,4,5, respectively (each kind of sliding window contains 64 filters) as \cite{kim2014convolutional}. Besides, we set the dimension of number embedding $d_n$ as 32, and the number of digit $l_n$ as 8, which meet $d_n \cdot l_n = d_c$.

In the training part, we set the learning rate of Adam optimizer as 0.001, and the dropout probability as 0.5. The batch size of all models is 128. We train every model for 16 epochs and evaluate the final model on the testing set.
\subsection{Results}
In this section, we compare our methods with existing methods and analyze the effort of each module in our methods.

To evaluate the performance of the proposed methods, we compare our methods with the following methods:
\begin{itemize}
\item \textbf{FLA:} a neural network method that capturing the interaction between Fact descriptions and applicable Laws with Attention mechanism. \cite{luo2017learning} .
\item \textbf{HARNN:} A Hierarchical Attention based RNN for document classification \cite{yang2016hierarchical}.
\item \textbf{CNN:} A CNN-based textual classification model \cite{kim2014convolutional}.
\item \textbf{TOPJUDGE:} A topological multi-task learning framework for LJP, which is the state-of-the-art method \cite{zhong2018legal}.
\end{itemize}

For each method in the baselines, we train each method by the multi-task way and select a set of the best experimental parameters according to the range of the parameters given in their experiments. We adopt $accuracy$ (Acc.), $macro$-$precision$ (MP), $macro$-$recall$ (MR) and $macro$-$F1$ (F1) which are widely used in the classification task to evaluate the performance of the baselines and our methods. Here, the macro-precision/recall/F1 are calculated by averaging the precision/recall/F1 of each category \cite{zhong2018legal}. All the models are repeated for 5 times, and we report the average values as the final results for clear illustration. 

Experimental results on the test sets of CAIL-small and CAIL-big are shown in Table \ref{cail-small} and Table \ref{cail-big}, respectively. The results show that MPBFN-WCA achieves the best performance on all metrics. Compared with TOPJUDGE, which performs best among baselines, MPBFN-WCA has enhanced the performance by about 2.75\%, 1.54\%, and 13.3\% on F1-score for law article prediction, charge prediction and term of penalty prediction in CAIL-small test set and about 2.24\%, 1.64\% and 8.92\% in CAIL-big test set. In other methods of baselines, the performance of CNN and HARNN are not as good as TOPJUDGE because they do not take the topological properties of multi-task into consideration. FLA is the worst performer because their work does not focus on the single-label problem. The performance of all methods in CAIL-big dataset is better than in CAIL-small, which is because the training data of CAIL-big is more adequate.

Besides, we also compare the performance of the different modules in our methods. In Table \ref{cail-small} and Table \ref{cail-big}, MPFP expresses the network of Multi-Perspective based Forward Prediction (no backward verification), MPBFN expresses the Multi-Perspective based Bi-Feedback Network, and MPBFN-WCA denotes MPBFN with WCA mechanism. Our MPBFN has a significant improvement in the term of penalty prediction. Even without WCA and BV, the penalty prediction result of MPFP is still better than baselines. We believe that this is caused by the multi-perspective mechanism since the penalty prediction has more than one pre-order task (it based on the result of law prediction and charge prediction). Compare with MPBFN, the WCA mechanism has enhanced the performance of term of penalty prediction by 4.81\% on F1-score in CAIL-small, and 3.11\% in CAIL-big. Even only employ MPBFN, the performance is still better than TOPJUDGE in two datasets, which indicates that the result-based multi-perspective bi-feedback network has a stable improvement in the performance of LJP.

\section{Conclusion}
In this paper, we focus on result-based multi-task topological dependencies of LJP. Specifically, based on the topology structure between multi-tasks, we design a network structure with multi-perspective forward prediction and backward verification to improve the performance of multitasking. Moreover, we extract the collocations in the fact description and combine the attention mechanism to reduce the misjudgment of penalty prediction. The experimental results show that our model achieves significant improvements over baselines on all judgment prediction tasks.

In the future, we will start with two aspects: (1)we explore the multi-task legal prediction with multi-label and multi-defendant; (2) we apply the multi-perspective bi-feedback network in other multi-task text classification problems.

\section*{Acknowledgments}
This work is supported by Chinese National Research Fund (NSFC) Key Project No. 61532013 and No. 61872239. FDCT/0007/2018/A1, DCT-MoST Joint-project No. (025/2015/AMJ), University of Macau Grant Nos: MYRG2018-00237-RTO, CPG2018-00032-FST and SRG2018-00111-FST of SAR Macau, China.

\bibliographystyle{named}
\bibliography{ijcai19}

\end{document}